\documentclass[sigconf]{acmart}

\usepackage{natbib}  
\usepackage{caption} 
\usepackage{subcaption}
\usepackage{enumitem} 
\usepackage{amsmath}
\usepackage{diagbox}
\usepackage{algorithmicx}
\usepackage{algorithm}
\usepackage{algpseudocode}
\AtBeginDocument{%
  \providecommand\BibTeX{{%
    \normalfont B\kern-0.5em{\scshape i\kern-0.25em b}\kern-0.8em\TeX}}}

\setcopyright{acmcopyright}
\copyrightyear{2018}
\acmYear{2018}
\acmDOI{}

\acmConference[Woodstock '18]{Woodstock '18: ACM Symposium on Neural
  Gaze Detection}{June 03--05, 2018}{Woodstock, NY}
\acmBooktitle{Woodstock '18: ACM Symposium on Neural Gaze Detection,
  June 03--05, 2018, Woodstock, NY}
\acmPrice{15.00}
\acmISBN{978-1-4503-XXXX-X/18/06}

\begin{document}

\title{KG-MTT-BERT: Knowledge Graph Enhanced BERT for Multi-Type Medical Text Classification}


\author{Yong He, Cheng Wang, Shun Zhang, Nan Li, Zhaorong Li, Zhenyu Zeng}
\email{{sanyuan.hy, youmiao.wc, changchuan.zs, kido.ln, 
zhaorong.lzr, zhenyu.zzy}@alibaba-inc.com}
\orcid{0000-0002-1585-2009,0000-0002-9554-2130,11:240000-0002-1679-6256,0000-0003-4430-7111,0000-0002-2047-6191,0000-0002-0329-6893}
\affiliation{ 
 \institution{Alibaba Group}
 \streetaddress{}
 \city{Hangzhou} 
 \country{China} 
 \postcode{}
}

\begin{abstract}
    Medical text learning has recently emerged as a promising area to improve healthcare due to the wide adoption of electronic health record (EHR) systems. The complexity of the medical text such as diverse length, mixed text types, and full of medical jargon, poses a great challenge for developing effective deep learning models. BERT has presented state-of-the-art results in many NLP tasks, such as text classification and question answering. However, the standalone BERT model cannot deal with the complexity of the medical text, especially the lengthy clinical notes. Herein, we develop a new model called KG-MTT-BERT (Knowledge Graph Enhanced Multi-Type Text BERT) by extending the BERT model for long and multi-type text with the integration of the medical knowledge graph. Our model can outperform all baselines and other state-of-the-art models in diagnosis-related group (DRG) classification, which requires comprehensive medical text for accurate classification. We also demonstrated that our model can effectively handle multi-type text and the integration of medical knowledge graph can significantly improve the performance.
\end{abstract}

\ccsdesc[500]{Computing methodologies~Language resources}
\ccsdesc[500]{Applied computing~Health informatics}

\keywords{EHR, BERT, Knowledge Graph, Multi-Type Medical Text,  Text Classification, Diagnosis-Related Group (DRG)}


\maketitle

\section{Introduction}

In recent years, the broad usage of electronic health record (EHR) systems has opened up many opportunities for applying deep learning methods to improve the quality of healthcare. Many predictive models have been developed for more effective clinical decision-making, such as predicting diagnoses and recommending medicine. With the rapidly aging population in many countries, the rational usage of scarce medical resources is critical, and developing a machine learning model for hospital resource management is crucial to improving healthcare quality and outcomes.

Diagnosis-related groups (DRGs) has been one of the most popular prospective payment systems introduced to put pressure on hospitals to optimize the allocation of medical resources in many countries. The basic idea is to classify inpatients into a limited number of DRG groups, assuming patients in the same DRG group share similar demographic and clinical characteristics and are likely to consume similar amounts of hospital resources. The reimbursement to the hospital is the same for the same DRG category, regardless of the variant actual costs for patients within the same DRG category. As a result, the DRG system can motivate hospitals to reduce costs, improve resource allocation, and increase operational efficiency.

The DRG used for reimbursement purposes is determined by two steps when a patient is discharged. A medical coder reviews the clinical documents, assigns standard medical codes, and selects one principal diagnosis following the guideline of coding and reporting using the International Classification of Diseases (ICD) or other classification systems. Then a DRG category is assigned with a set of rules implemented in a software (i.e. DRG grouper) using the following variables: principal diagnosis code, secondary diagnosis codes, procedure codes, age, sex, discharge status, and length of stay. In addition to post-hospital billing, DRG can be repurposed for in-hospital operation management and a previous study has shown that classifying a patient’s DRG during the hospital visit can allow the hospital to more effectively allocated hospital resources and to improve the facilitation of the operational planning \citep{drg}. Directly inferring DRG from the complex medical text of EHR is a challenge because the model needs to mimic the professional medical coder’s expertise and learn the rule of the DRG grouper.

Multiple types of medical texts from EHR are closely related to DRG, such as diagnosis, procedure, and hospital course. We formulate the DRG classification task as a multi-type text classification problem. The following characteristic of these texts poses a great challenge for medical text mining. 1) Diverse length: diagnoses and procedures are typically short texts, while the hospital course is always lengthy with detailed information of current hospital visit; 2) Multiple Types: the different fields in EHR are to serve different purposes and thus texts in different fields are of different types (See Appendix 
A.1 for more details of multi-type medical text); 3) Medical-domain specific: the free text from EHR contains a lot of medical terminology and a medical coder assigns diagnosis code with relevant domain knowledge.

BERT\citep{bert} has been widely used in text classification and typically multiple texts are concatenated into one long text for modeling purposes. However, there are three limitations of directly applying the BERT model for multi-type texts. First, the input of BERT is limited to 512 tokens, and combining all texts could easily exceed this limit and lead to information loss. Second, the concatenation of multiple types of text may not be feasible as some texts are syntactically different or contextually irrelevant, which makes the direct self-attention module of BERT unnecessary. Lastly, BERT was pre-trained over open-domain corpora and it does not play well with EHR data in the medical field.

To overcome the above limitations for multi-type text classification, we develop a new model KG-MTT-BERT (Knowledge Graph Enhanced Multi-Type Text BERT) to extend the BERT model for multi-type text and integrate medical knowledge graph for domain specificity. Our model first applies the same BERT-Encoder to process each text, i.e. the encoder of each text of one patient shares the same parameters. The two levels of encoding outputs from BERT-Encoder with different granularities, one is at text-level and the other is at token-level, are compared in this paper. Multiple encodings from input texts are concatenated together as the representation matrix and different types of pooling layers are investigated for better information summation. In addition, we use the knowledge graph to enhance the model’s ability to handle the medical domain knowledge. Finally, a classification layer is utilized to predict DRG categories. 

We have conducted experiments on three DRG datasets from Tertiary hospitals in China (See Appendix A.2 for the definition of hospital classification). Our model can outperform baselines and other state-of-the-art models in all datasets. The rest of the paper is organized as follows. We review related works and techniques in the Related Works Section and introduce the formulation and architecture of our model in the Method Section. Then we report the performances of our model in three datasets and investigate hyperparameter effects and multiple ablation studies to further the understanding of the model. Lastly, we conclude our paper and point out our future directions.

\section{Related Works}
Text classification is an important Natural Language Processing (NLP) task directly related to many real-world applications. The goal of text classification is to automatically classify the text into predefined labels. The unstructured nature of free text requires the transformation of text into a numerical representation for modeling. Over the past decade, text classification has changed from shallow to deep learning methods, according to the model structures \citep{li2020survey}. Shallow learning models focus on the feature extraction and classifier selection, while deep learning models can perform automatic high-level feature engineering and then fit with the classifier together. 

Shallow learning model first converts text into vectors using text representation methods like Bag-of-words (BOW), N-gram, term frequency-inverse document frequency (TF-IDF) \citep{tf_idf}, Word2vec \citep{word2vec}, and GloVe \citep{glove}, then trains the shadow model to classify, such as Naive Bayes \citep{nb}, Support Vector Machine \citep{svm}, Random Forest \citep{rf}, XGBoost \citep{xgboost}, and LightGBM \citep{lightgbm}. In practice, the classifier is trained and routinely selected from the zoo of shallow models. Therefore, feature extraction and engineering is the critical step for the performance of text classification. 

Various deep learning models have been proposed in recent years for text classification, which builds on basic deep learning techniques like CNN, RNN, and the attention mechanism. TextCNN \citep{textcnn} applies the convolutional neural network for sentence classification tasks. The RNNs, such as long short-term memory (LSTM), are broadly used to capture long-range dependence. Zhang et al. introduce bidirectional long short-term memory networks (BLSTM) to model the sequential information about all words before and after it \citep{textrnn}. Zhou et al. integrate attention with BLSTM as Att-BLSTM to capture the most important semantic information in a text \citep{textrnn_at}. Recurrent Convolutional Neural Network (RCNN) combines recurrent structure to capture the contextual information and max-pooling layer to judge the key message from features \citep{textrcnn}. Therefore, RCNN leverages the advantages of both RNN and CNN. Johnson et al. develop a deep pyramid CNN (DPCNN) model to increase the depth of CNN by alternating a convolution block and a down-sampling layer over and over \citep{dpcnn}. Yang et al. design a hierarchical attention network (HAN) for document classification by aggregating important words to a sentence and then aggregating importance sentences to document \citep{han}. The appearance of BERT, which uses the masked language model to pre-train deep bidirectional representation, is a significant turning point in developing text classification models and other NLP technologies. The memory and computation cost of self-attention grows quadratically with text length and prevents applications for long sequences. Longformer \citep{longformer} and Reformer \citep{reformer} are designed to address this limitation.

Domain knowledge serves a principal role in many industries. In order to integrate domain knowledge, it is necessary to embed the entities and relationships of the knowledge graph. Many knowledge graph embedding methods have been proposed, such as TransE \citep{TransE}, TransH \citep{TransH}, TransR \citep{TransR}, TransD \citep{TransD}, KG2E \citep{KG2E}, and TranSparse \citep{TranSparse}. Some existing works show that the knowledge graph can improve the ability of BERT, such as K-BERT \citep{K-BERT} and E-BERT \citep{E-BERT}.

Data mining of EHR \citep{emr_mining} and the biomedical data has been an increasingly popular research area, such as medical code assignment, medical text classification \citep{mdeical_text_classification}, and medical event sequential learning. Several models have been developed for medical code assignment, which is a prerequisite of the DRG grouper. Multimodal Fusion Architecture SeArch (MUFASA) model investigates multimodal fusion strategies to predict the diagnosis code from comprehensive EHR data \citep{xu2021mufasa}. Yan et al. design a knowledge-driven generation model for short procedure mention normalization \citep{yan2020knowledge}. Limited research has been performed on direct DRG prediction. Gartner et al. apply the shallow learning approach to predict the DRG from the free text with complicated feature engineering, feature selection, and 9 classification techniques \citep{drg}. AMANet \citep{amanet} treats the DRG classification as a dual-view sequential learning task based on standardized diagnosis and procedure sequences. BioBERT \citep{biobert} is a domain-specific language representation model pre-trained on large-scale biomedical corpora, which is based on BERT. Med-BERT \citep{medbert} adapts the BERT framework for pre-training contextualized embedding models on the structured diagnosis ICD code from EHR. ClinicalBERT \citep{ClinicalBERT} uses clinical text to train the BERT framework for predicting 30-day hospital readmission at various time points of admission.

\section{Method}
\subsection{Problem Formulation}
In this section, we define the notations used in this paper. The goal of the text classification task is to predict the target $Y$ given multi-type input texts $Xs$, where $s$ stands for the type id. Let $S_i=\{Xs_i, T_i, y_i\}$ to be the sample with index $i$, where $Xs_i=\{X_{i1}, X_{i2}, ..., X_{ik} \}$ is the input texts of sample $i$, $X_{ij}=\{x_{ij1}, x_{ij2}, ..., x_{ijl_{ij}}\}$ is the $j$-th text of sample $i$, $k$ is the number of input text types, $l_{ij}$ is the length of the text $X_{ij}$, $x_{ijp}$ is the $p$-th token of the text $X_{ij}$, $T_i=\{t_{i1}, t_{i2},...,t_{im}\}$ is the relevant entities of sample $i$, and $y_i$ is the target label. The entities, including the principal diagnosis, principal procedure, and symptoms, are extracted from input texts. 

In the DRG classification task, we predict the DRG according to basic information, chief complaint, history of present illness, hospital course, procedure text, diagnosis text, and the entities extracted from the patient’s medical record of the current visit. This is a multi-class classification task with $d_y$ categories.

\subsection{Medical Knowledge Graph} 
Our model integrates the knowledge graph to fine-tune the Chinese medicine clinical text classification task. And the knowledge graph is dynamically built from the training set in each dataset by extracting entities and the co-occurrence relations between these entities. We build the Knowledge Graph for each dataset independently using entities extracted from each \textbf{training set}. The medical knowledge graph includes four types of entities, i.e., DRG, diagnosis, procedure, and symptom, and according to the meaning of these co-occurrence relations, only four of them are left, i.e., DRG-related diagnosis, DRG-related procedure, diagnosis-related procedure, and diagnosis-related symptom. The symptom set is from the Chinese medial knowledge database OMAHA \footnote[1]{https://www.omaha.org.cn/}. The Chinese names of diagnosis and procedure are from ICD-10 and ICD-9-CM-3 in China, respectively. The DRG comes from multiple different versions. We extract the entity from the medical text by going through the NER (named-entity recognition) and the subsequence matching algorithm, and these extracted entities are matched to the standard entity names through the string similarity measure algorithm based on the LCS (longest common subsequence). The process is shown in Figure 1. The symptoms are extracted from the EHR field chief complaint and history of present illness. The principal diagnosis and principal procedure entities are extracted from EHR field diagnosis and procedure (free text), respectively. DRG is the category label. The extracted entities are linked if they are in the same hospital visit. More information on medical knowledge graph building and statistics can be found in the Appendix Medical Knowledge Graph Section. A part of the knowledge graph is shown in Figure 2. 
\begin{figure}[ht]
  \centering
  \label{extracting_names}
  \includegraphics[width=1\columnwidth]{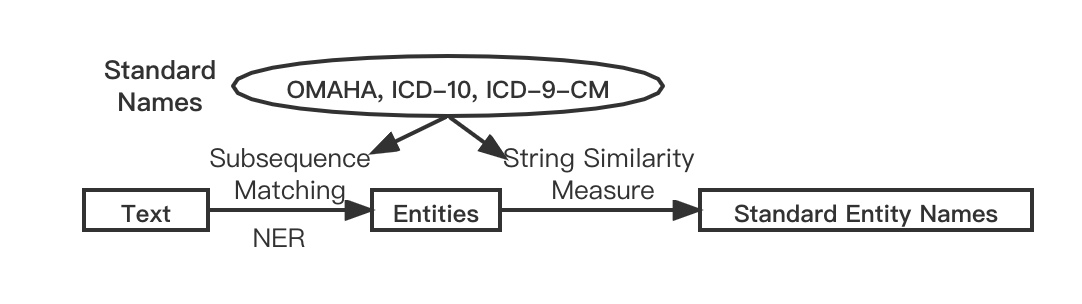}
  \caption{Extracting standard entity names from text. }
\end{figure}

\begin{figure}[ht]
  \centering
  \label{part_kg}
  \includegraphics[width=1\columnwidth]{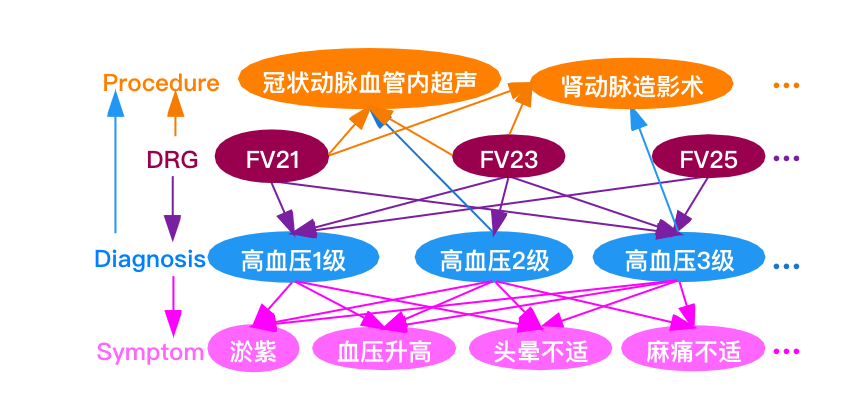}
  \caption{A part of the medical knowledge graph. }
\end{figure}

\subsection{Network Architect}
Our model is named KG-MTT-BERT (Knowledge Graph Enhanced Multi-Type Text BERT), and its structure is shown in Figure 3. The pseudo-code of our model is in Appendix Pseudo-code Section. We will open our code to GitHub. In this section, we describe the modules of our model in detail.

\begin{figure*}[ht]
     \centering
     \begin{subfigure}[b]{0.49\textwidth}
         \centering
         \includegraphics[width=\textwidth]{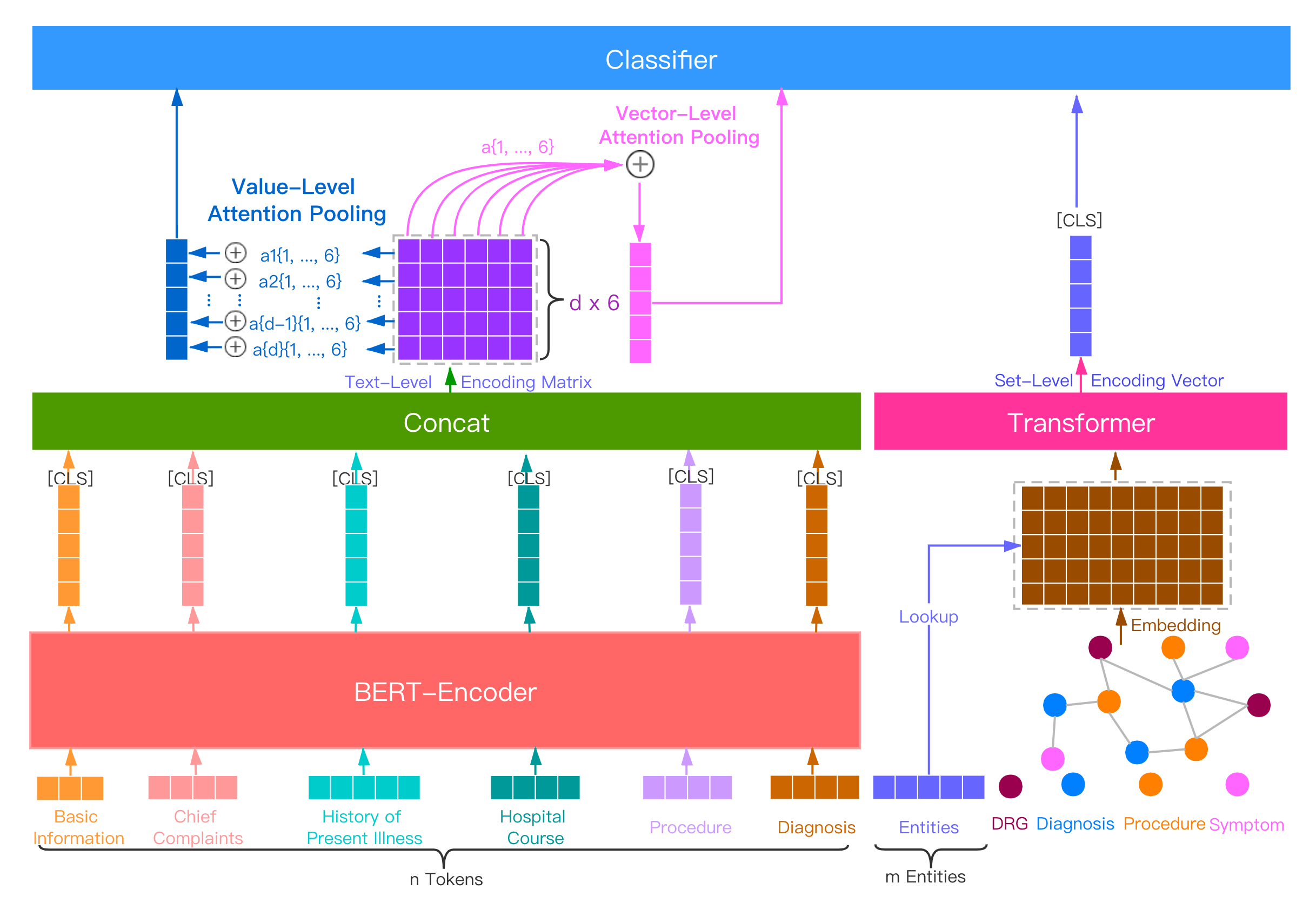}
         \caption{Text-Level Encoding with Vector-Level and Value-Level Attention Pooling.}
         \label{fig:Text-Level Encoding }
     \end{subfigure}
     \hfill
    \begin{subfigure}[b]{0.49\textwidth}
         \centering
         \includegraphics[width=\textwidth]{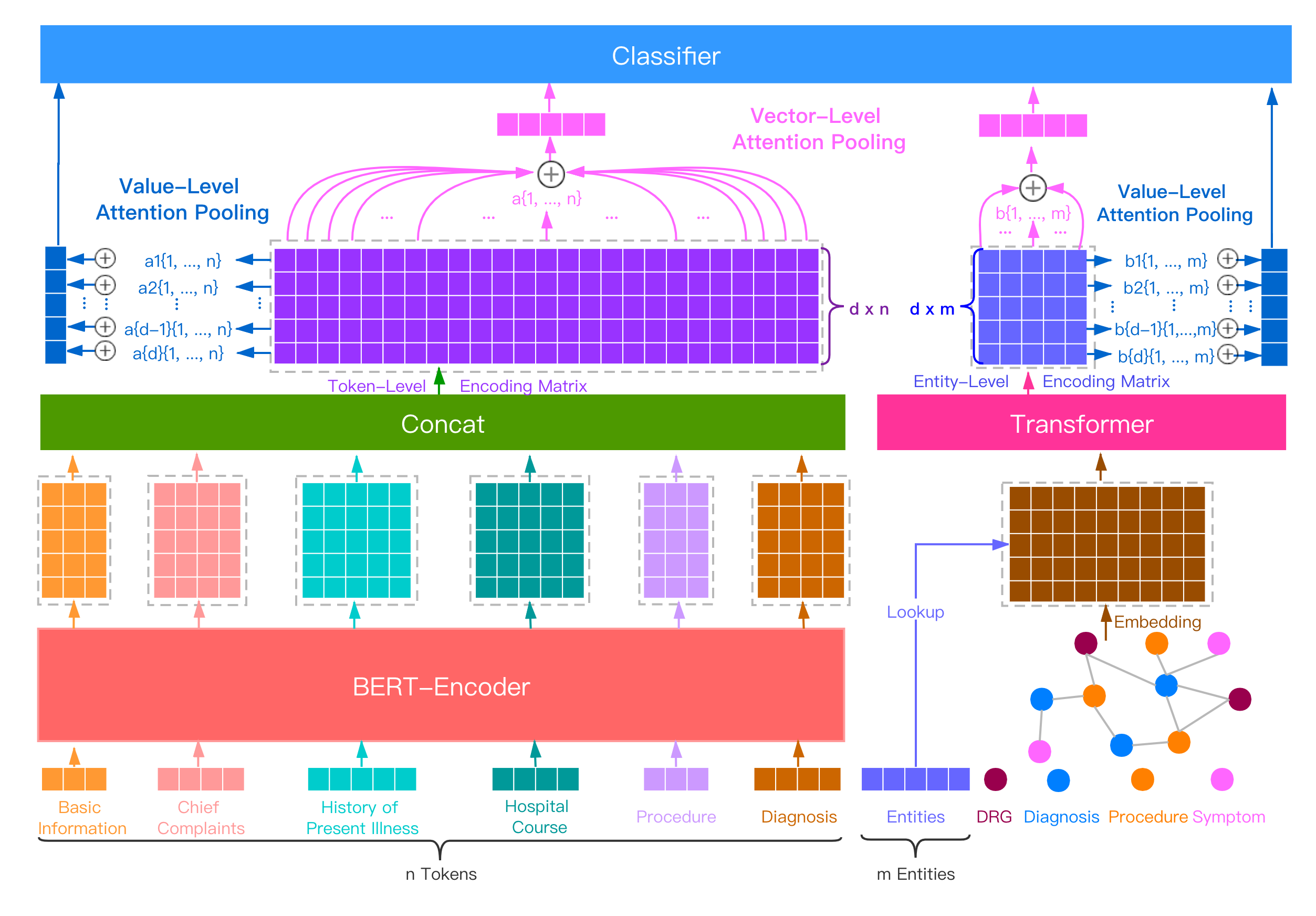}
         \caption{Token-Level Encoding with Vector-Level and Value-Level Attention Pooling.}
         \label{fig:Token-Level Encoding 1}
     \end{subfigure}
    \caption{KG-MTT-BERT for Six Types of Medical Texts.}
    \label{fig:our model}
\end{figure*}

\paragraph{\textbf{BERT Encoder}} 
We use the same BERT Encoder to extract the features of the six types of input texts respectively to solve the problem that a single text generally doesn't exceed 512 but the six types of texts are combined and exceed the length limit. And The encoder output two types of encoding vectors: text-level encoding vectors and token-level encoding vectors. \\ 
\textbf{Text-Level Encoding.}  
It saves the last layer hidden state vector $h^{[CLS]}$ of the first token [CLS] for the text-level representation of each text. 
\begin{equation}
\begin{aligned}
E_i &= \textrm{Encoder}(Xs_i) = \textrm{Encoder}(\{X_{i1}, ..., X_{ik} \}) \\ 
    &= \{\{h_{i1}^{[CLS]}\}, ..., \{ h_{ik}^{[CLS]}\}\}
\end{aligned}
\end{equation}
where $i$ is the sample id, $k$ is the number of input text types. \\ 
\textbf{Token-Level Encoding.} It saves the last layer hidden state matrix $H$ of each text as the token-level encoding. The vector $h_{ijp}$ of $H_{ij}$ is the encoding vector of token $x_{ijp}$ of text $X_{ij}$ of sample ${i}$. 
\begin{equation}
\begin{aligned}
E_i &= \textrm{Encoder}(Xs_i) = \textrm{Encoder}(\{X_{i1}, ..., X_{ik} \}) \\ &= \{\{H_{i1}\}, ..., \{H_{ik}\} \} \\ &=  \{\{h_{i11}, .., h_{i1l_{i1}}\}, ..., \{h_{ik1}, ..., h_{ikl_{ik}} \} \}
\end{aligned}
\end{equation}
where $i$ is the sample id, $k$ is the number of input text types, $l_{ik}$ is the length of text $X_{ik}$.

Limited by machine resources and to simplify the model,  all types of input texts share the same BERT-Encoder.

\paragraph{\textbf{Concatenate Layer}} 
This layer concatenates the encoded vectors or matrices $E_i$ into matrix $C_i$: 
\begin{equation}
\begin{aligned}
C_i &= \{h_{i1}^{[CLS]}, h_{i2}^{[CLS]}, ..., h_{ik}^{[CLS]}\}
\end{aligned}
\end{equation}
\begin{equation}
\begin{aligned}
C_i &= \{h_{i11}, .., h_{i1l_{i1}}, ..., h_{ik1}, ..., h_{ikl_{ik}} \}
\end{aligned}
\end{equation}

\paragraph{\textbf{Pooling Layer}} 
This layer transforms the encoded matrix $C_i$ to the final representation vector $e_i$ of all input texts. We introduce two types of pooling methods here. \\
\textbf{Vector-Level Attention Pooling.} This method calculates the weight $\alpha$ for each vertical vector of encoding matrix $C_i$. The representation vectors for text-level encoding and  are calculated as follows: 
\begin{equation}
\begin{aligned}
e_i = \sum_{j=1}^{k} \alpha_{ij} h_{ij}^{[CLS]}, \sum_{j=1}^{k} \alpha_{ij}=1 
\end{aligned}
\end{equation}
\begin{equation}
\begin{aligned}
e_i = \sum_{j=1}^{k} \sum_{p=1}^{l_{ij}} \alpha_{ijp} h_{ijp}, \sum_{j=1}^{k} \sum_{p=1}^{l_{ij}} \alpha_{ijp} = 1
\end{aligned}
\end{equation} \\ 
\textbf{Value-Level Attention Pooling.}
This method calculates the weight for each value in the encoding matrix $C_i$, which is different from vector-level attention pooling where the values in each vertical vector share the same weight. The representation vectors for text-level encoding and token-level encoding are calculated as follows:
\begin{equation}
\begin{aligned}
e_{iq} &= \sum_{j=1}^{k} \alpha_{ijq} h_{ijq}^{[CLS]}, \sum_{j=1}^{k} \alpha_{ijq} = 1
\end{aligned}
\end{equation}
\begin{equation}
\begin{aligned}
e_{iq} &= \sum_{j=1}^{k} \sum_{p=1}^{l_{ij}} \alpha_{ijpq} h_{ijpq}, \sum_{j=1}^{k} \sum_{p=1}^{l_{ij}} \alpha_{ijpq} = 1
\end{aligned}
\end{equation}
\begin{equation}
\begin{aligned}
e_i &= [e_{i1}, e_{i2}, ..., e_{id}]
\end{aligned}
\end{equation}
where $d$ is the dimension of encoding vector, and $q \in \{1,2, ..., d\}$.

Four types of pooling strategies are investigated in this research to determine the weight $\alpha$ in Eq. (5) to Eq. (8). \\  
\textbf{Global Max Pooling.} This a value-level hard pooling strategy, which takes the maximum value of each horizontal dimension of the encoded vectors. And the representation vectors $e_{i}$ of text-level and token-level encoding are as follows: 
\begin{equation}
\begin{aligned}
e_i &= \max\{h_{i1}^{[CLS]}\, h_{i2}^{[CLS]}, ..., h_{ik}^{[CLS]}\}
\end{aligned}
\end{equation}
\begin{equation}
\begin{aligned}
e_i &= \max\{h_{i11}, ..., h_{i1l_{i1}}, ...,  h_{ik1}, ..., h_{ikl_{ik}}\}
\end{aligned}
\end{equation}
where $\max$ operation calculates the max value across each dimension of the input vectors. \\ 
\textbf{Contextual Attention Pooling.} It computes the weight $\alpha$ in Eq. (5) to Eq. (8) similar to the word attention and sentence attention in the HAN model \citep{han}, which introduces the context vector in attention weight calculation. For vector-level contextual attention pooling, the weights $\alpha_{ij}$ in Eq. (5) and $\alpha_{ijp}$ in Eq. (6) are calculated as follows: 
\begin{equation}
\begin{aligned}
\alpha_{ij} = e^{\mu_{ij}^T w}/\sum_{t}e^{\mu_{it}^T w}, \mu_{ij} = \tanh(U h_{ij}^{[CLS]} + b)
\end{aligned}
\end{equation}
\begin{equation}
\begin{aligned}
\alpha_{ijp} = e^{\mu_{ijp}^T w}/\sum_{t_1}\sum_{t_2}e^{\mu_{it_1t_2}^T w}, \mu_{ijp} = \tanh(U h_{ijp} + b)
\end{aligned}
\end{equation}
where $U$ is a trainable matrix, $b$ is a bias trainable vector, and $w$ is a trainable context vector. 

The value-level contextual attention pooling is calculated similarly, and the weights $\alpha_{ijq}$ in Eq. (7) and $\alpha_{ijqp}$ in Eq. (8) are calculated as follows:
\begin{equation}
\begin{aligned}
\alpha_{ijq} = \frac{e^{\mu_{ij}^T w_q}}{\sum_{t}e^{\mu_{it}^T w_q}}, \mu_{ij} = \tanh(U h_{ij}^{[CLS]} + b) 
\end{aligned}
\end{equation}
\begin{equation}
\begin{aligned}
\alpha_{ijpq} = \frac{e^{\mu_{ijp}^T w_q}}{\sum_{t_1}\sum_{t_2}e^{\mu_{it_1t_2}^T w_q}}, \mu_{ijp} = \tanh(U h_{ijp} + b)
\end{aligned}
\end{equation}
where $q \in \{1,2, ..., d\}$, $d$ is the dimension of encoding vector, $W$ is a trainable context matrix with $d$ columns, and $w_q$ is the $q$-th column of matrix $W$.   \\ 
\textbf{Global Average Pooling.} This is a special pooling strategy with all weights $\alpha$ are equal. Therefore, the representation vectors $e_{i}$ of text-level and token-level encoding are as follows: 
\begin{equation}
\begin{aligned}
e_i &= \frac{\sum_{j=1}^{k} h_{ij}^{[CLS]}}{k}
\end{aligned}
\end{equation}
\begin{equation}
\begin{aligned}
e_i &= \frac{\sum_{j=1}^{k}\sum_{p=1}^{l_{ij}} h_{ijp}}{\sum_{j=1}^{k} l_{ij}}
\end{aligned}
\end{equation}
\textbf{Transformer Pooling.} It applies the self-attention mechanism relating different positions of the encoded matrix $C_i$ to compute a representation vector. It can capture the relationships among vectors and determine the weights. First, insert a trainable vector $c_0$ into the first position of $C_i$, $c_0$ can be regarded as a new [CLS] embedding vector.
\begin{equation}
\begin{aligned}
C_i'=\{c_0, c_{i1}, c_{i2}, ...\}
\end{aligned}
\end{equation}

Then a transformer layer with two encoder blocks is applied to obtain $e_i$.
\begin{equation}
\begin{aligned}
e_i = \textrm{transformer}(C_i')
\end{aligned}
\end{equation}

\paragraph{\textbf{Knowledge Graph Enhanced Module}}   
The clinical text contains more medical terms, and the text processing granularity of BERT is at the token level, so the knowledge graph enhanced module is designed to enhance the importance of entities in these texts and to integrate domain knowledge into the model. Firstly, we use KG2E \citep{KG2E} to embed the entities of the knowledge graph to obtain the matrix $E^{KG}$, which is used as the initial embedding matrix of all entities. The input entities $T_i=\{t_{i1}, t_{i2}, ..., t_{im}\}$ of each sample are transformed to embedding matrix $E^{KG}_i$ by looking up the embedding table. A transformer layer with one encoder block and no positional encoding is applied to the embedded matrix $E^{KG}_i$ for feature extraction and two types of encoding outputs, i.e., entity-level encoding (entities encoding matrix $M^{KG}_i$) and set-level encoding ([CLS] encoding vector $e^{KG}_i$), are saved similar to the feature extraction of texts. For entity-level encoding, the vector-level attention pooling or the value-level attention pooling mentioned above is used to transform matrix $M^{KG}_i$ into vector $e^{KG}_i$.
\begin{equation}
\begin{aligned}
e^{KG}_i = \textrm{transformer}(\{[CLS],t_{i1},t_{i2},...,t_{im}\})=h_i^{[CLS] KG}
\end{aligned}
\end{equation}
\begin{equation}
\begin{aligned}
e^{KG}_i &= \textrm{pool}(\textrm{transformer}(\{t_{i1},t_{i2},...,t_{im}\})) \\ 
    &= \textrm{pool}(\{h_{i1}^{KG},h_{i2}^{KG},...,h_{im}^{KG}\})=\textrm{pool}(M^{KG}_i)
\end{aligned}
\end{equation}

\paragraph{\textbf{Classification Layer}}   
Use the text representation vector $e_i$ and the knowledge graph enhanced vector $e^{KG}_i$ as the inputs for final classification as:
\begin{equation}
\begin{aligned}
\hat{y_i} = \textrm{softmax}(\textrm{concat}(e_i, e^{KG}_i))
\end{aligned}
\end{equation}

\subsection{Loss Function}   
We use the cross-entropy loss for this multi-class classification task.

\section{Experiments}     
In this section, we first introduce three DRG datasets and their statistics. Then, we compare our model to baselines and other state-of-the-art models in these datasets. Finally, we evaluate the performance of our model in different hyperparameter settings, encoding methods, pooling strategies, and with or without the knowledge graph.

\subsection{Datasets}  
Due to the inconsistent medical informatization level of different regions, there are five more versions of DRG groupers co-existing in China. Three Chinese DRG datasets\footnote[2]{The approval process for opening the three datasets is in progress.}, representing the results of three popular versions of DRG groupers, are from three tertiary hospitals in different provinces of China and we label them as DRG-A, DRG-B, and DRG-C. All of these datasets are desensitized and licensed by the data provider. In all datasets, each sample consists of six types of Chinese texts, extracted Chinese entities, and the DRG label. These Chinese texts include basic information, chief complaint, history of present illness, hospital course, diagnosis, and procedure (See Appendix A.1 for more details of multi-type medical text). The basic information consists of sex, age, the hospital department, and hospitalization days of a hospital visit. The chief complaint is a medical term to describe the primary concern of the patient that led the patient to seek medical advice. The history of present illness is a description of the development of the patient's present illness. The hospital course records detailed information about a sequence of events from a patient’s admission to discharge during the current visit. The complete set of a patient's diagnoses from the doctor are combined as a diagnosis text, and all procedures are combined as a procedure text. The entities, including principal diagnosis, principal procedure, and symptoms, are extracted from these texts. The DRG label is determined by the DRG grouper for classification. The statistics of the three datasets are summarized in Table 1. 

\begin{table*}[htbp]
  \label{Three Datasets Label}
  \small
  \centering
  \caption{Three Datasets}
  \begin{tabular}{l|lll|lll|lll}
       \hline 
       {Dataset}&\multicolumn{3}{c|}{DRG-A}&\multicolumn{3}{c|}{DRG-B}&\multicolumn{3}{c}{DRG-C} \\
       \hline
       {}&\multicolumn{3}{c|}{Count}&\multicolumn{3}{c|}{Count}&\multicolumn{3}{c}{Count} \\
       \hline 
       {Samples}&\multicolumn{3}{c|}{15407}&\multicolumn{3}{c|}{36528}&\multicolumn{3}{c}{8434} \\
       {Labels}&\multicolumn{3}{c|}{177}&\multicolumn{3}{c|}{317}&\multicolumn{3}{c}{109} \\
	   \hline 
	   \diagbox{Input Type}{ Length}&{Min}&{Max}&{Mean}&{Min}&{Max}&{Mean}&{Min}&{Max}&{Mean}  \\
	   \hline 
	   Chief Complaint  & 5 & 41 & 13.2 & 5 & 48 & 14.5 & 3 & 32 & 13.0 \\
       History of Present Illness & 11 & 1517 & 267.0 & 7 & 4236 & 335.6 & 77 & 1034 & 270.0 \\
       Hospital Course &  34 & 5417 & 591.0 & 12 & 2135 & 139.0 & 12 & 5701 & 644.1 \\
       Basic Information & 9 & 18 & 12.9 & 12 & 19 & 15.0 & 9 & 16 & 11.5 \\
       Diagnosis  & 3 & 804 & 34.3 & 3 & 216 & 42.9 & 3 & 187 & 32.9 \\
       Procedure  & 2 & 192 & 12.6 & 2 & 190 & 13.4 & 2 & 134 & 8.2 \\
       Entities  & 1 & 11 & 2.4 & 1 & 14 & 2.1 & 1 & 10 & 2.8 \\
	   \hline 
  \end{tabular}
\end{table*}

\subsection{DRG Classification}
\paragraph{\textbf{Baselines}} We compare our model with the following baselines and state-of-the-art algorithms. It is worth noting that \textbf{BioBERT} is pre-trained on English biomedical corpora, \textbf{ClinicalBERT} is pre-trained on English clinical text, and \textbf{Med-BERT} is pre-trained on the standard diagnostic ICD code from patient EHR data. But our task uses the Chinese clinical texts and the Chinese Medical Knowledge Graph to perform DRG classification. Therefore, these three models are not suitable for our tasks, and the pre-trained \textbf{``BERT-Chinese-base''} model is used as the basis of our model.

\textbf{FastText\citep{fasttext}.} It is an open-source, free, lightweight library for efficient learning of word representations and text classification.

\textbf{TextCNN\citep{textcnn}.} It employs CNN to automatically extract features from the text. 

\textbf{Att-BLSTM\citep{textrnn_at}.} It integrates BLSTM and attention to assign the contribution of each word for text classification. 

\textbf{RCNN\citep{textrcnn}.} It applies a recurrent structure and a max-pooling layer to automatically judge which words play key roles in text classification. 

\textbf{DPCNN\citep{dpcnn}.} This is a pyramid model with an increased depth of the CNN block for long-range contextual information. 

\textbf{HAN\citep{han}.} HAN model builds a document vector for classification by progressively aggregating important words into sentence vectors and then aggregating critical sentence vectors to document vector.  

\textbf{Reformer\citep{reformer}.} It replaces self-attention with locality-sensitive hashing (LSH) and local attention to reduce the complexity of computation.

\textbf{Lonformer\citep{longformer}.} It proposes an attention mechanism combining local and global information to scale linearly with the sequence length.

\textbf{BERT\citep{bert}.} This is the most popular pre-trained model nowadays and we fine-tune \textbf{``BERT-Chinese-base''} for text classification.

\paragraph{\textbf{Text Combination}}
The different types of texts of each sample should be concatenated together as one input text for modeling in FastText, TextCNN, Att-BLSTM, RCNN, DPCNN, and BERT. Based on the intuition, three types of texts, i.e., chief complaint, procedure, and diagnosis, contribute more to the DRG assignment than the rest of the three types of texts. Therefore, the input texts are concatenated as one long text in the order of chief complaint, procedure, diagnosis, hospital course, history of present illness, and basic information. There is no requirement for a text combination of HAN and our model. We treat each type of text as a sentence in the HAN model. For BERT, the long text is truncated to a maximum of 510 tokens. Our model can support up to 510 tokens for each text type. In BERT and our model, [CLS] and [SEP] are added to the beginning and end of the text respectively to form a maximum of 512 tokens.

\paragraph{\textbf{Evaluation Metrics}} We use accuracy, F1-score, and the area under the Precision-Recall curve (PRAUC) as the evaluation metrics for this task, where F1-score and PRAUC are macro-averaged. We randomly divide the dataset into training, validation, and test set with a ratio of 7:1.5:1.5 and report the average performance of 5 runs from the test set.

\paragraph{\textbf{Implementation Details}} The pre-trained \textbf{``BERT-Chinese-base''} model (L=12, H=768, A=12) has been used in all models for a different purpose. Our model uses its parameter values as initial weights of the encoder layer. Other models use its token embedding as an initial embedding matrix. Among the best results of all algorithms, the initial learning rates of the evaluated models are 0.00001 for BERT, Reformer, Longformer, and our model, 0.002 for DPCNN, and 0.001 for the rest of the models. The batch size of the training is 8. The training epoch is 30 for our model, which converges between the 20th and 30th epoch, and is 50 for other models. In Figure 4, we show the loss curve of the training set over the training epochs on the DRG-B dataset, as well as the accuracy of the validation set over training epochs. All experiments are performed using GeForce RTX 3090 in the Ubuntu 18.04.1 system. See Appendix Execution Information Section for more details of hyperparameter search for each model, especially for Reformer and Longformer.

\begin{figure*}
     \centering
     \begin{subfigure}[b]{0.49\textwidth}
         \centering
         \includegraphics[width=\textwidth]{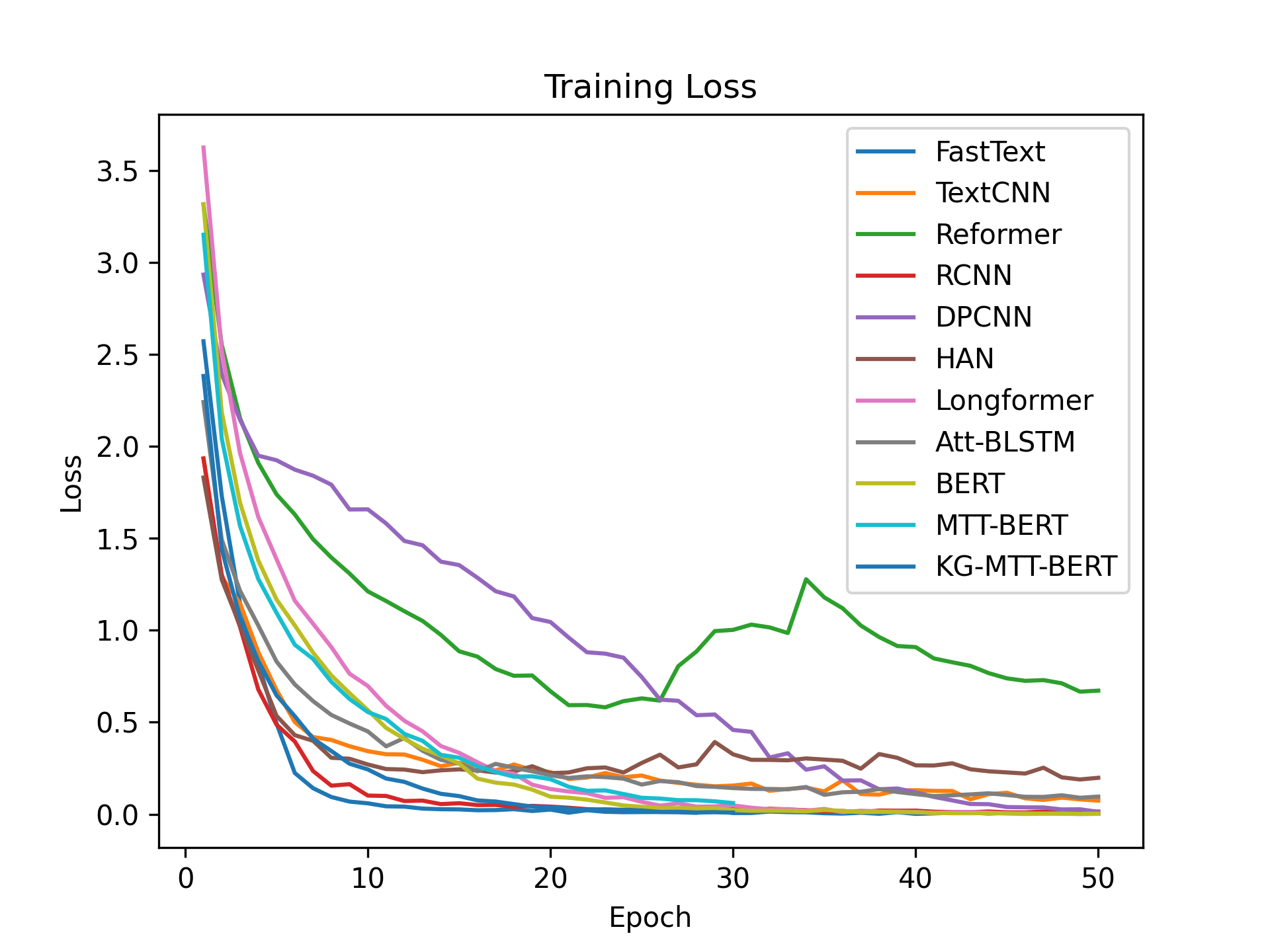}
         \subcaption*{Training Loss}
         \label{fig:Training Loss}
     \end{subfigure}
     \begin{subfigure}[b]{0.49\textwidth}
         \centering
         \includegraphics[width=\textwidth]{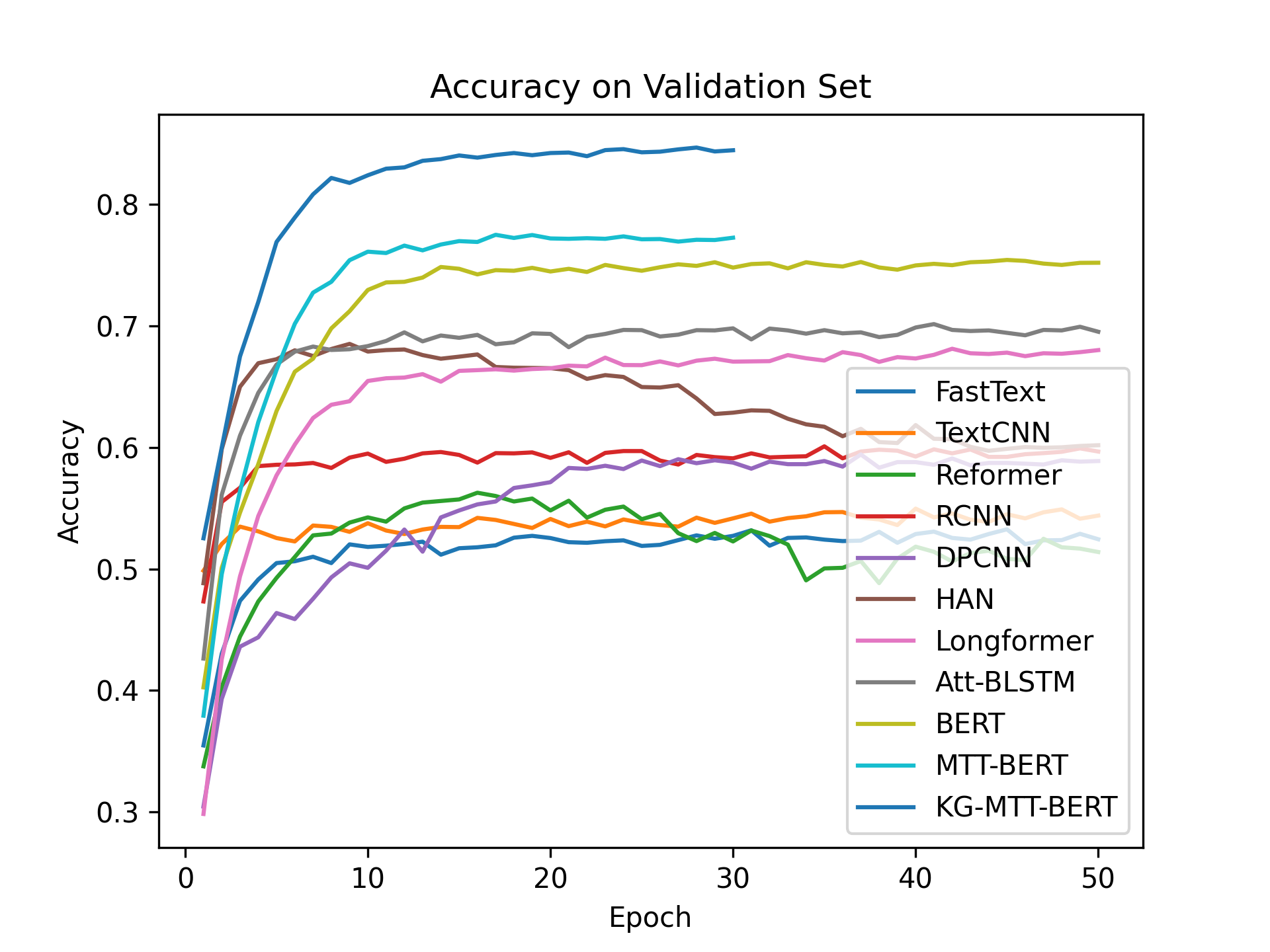}
         \subcaption*{Accuracy on Validation Set}
         \label{fig:Accuracy on Validation Set}
     \end{subfigure}
    \caption{Curves of loss on the training set and accuracy on the validation set VS. training epochs on DRG-B dataset. The training epoch is 30 for our model and is 50 for other models.}
    \label{fig:curves}
\end{figure*}

\paragraph{\textbf{Results}} Given the different combination of encoding method and pooling strategy proposed in the method section, our final model is token-level encoding with global max pooling and the details of the ablation study are shown in the ablation study section. Table 2, Table 3, and Table 4 summarize our results in three datasets and we can see that our model outperforms other models in all metrics. Our model is improved after the integration of the knowledge graph. For example, the PRAUC is improved from 0.8741 to 0.9310 with the knowledge graph in the DRG-A dataset. Comparing to BERT, our model can consistently achieve significantly better results. We can observe the improvement of 0.04 to 0.08 in accuracy, 0.03 to 0.07 in F1, and 0.07 to 0.13 in PRAUC over three datasets. 

\begin{table}
  \label{Experiments on DRG-A Dataset}
  \centering
  \small
  \caption{Experiments on DRG-A Dataset}
  \begin{tabular}{llll}
    \hline
    Method & Acc &F1 &PRAUC \\
    \hline 
	FastText&0.6428&0.5760&0.6295 \\
	TextCNN&0.7214&0.6570&0.6836 \\
	Reformer&0.7197&0.6621&0.7306 \\
	DPCNN&0.7788&0.7261&0.7642 \\
    HAN&0.8048&0.7616&0.8018 \\
    Longformer&0.8175&0.7794&0.7862 \\
	Att-BLSTM&0.8185&0.7780&0.8109 \\
	RCNN&0.8287&0.7887&0.8293 \\
	BERT&0.8541&0.8189&0.8552 \\
	\hline  
    \textbf{MTT-BERT(w/o KG)}&0.8703&0.8425&0.8741 \\
    \textbf{KG-MTT-BERT}&\textbf{0.9179}&\textbf{0.8961}&\textbf{0.9310} \\
    \hline
  \end{tabular}
\end{table}

\begin{table}
  \label{Experiments on DRG-B Dataset}
  \centering
  \small
  \caption{Experiments on DRG-B Dataset}
  \begin{tabular}{llll}
    \hline
    Method & Acc &F1 &PRAUC \\
    \hline  
	FastText&0.5235&0.4732&0.4866 \\
	TextCNN&0.5306&0.4824&0.4662 \\
	Reformer&0.5553&0.5008&0.5284 \\
	RCNN&0.5930&0.5526&0.5921 \\
	DPCNN&0.5985&0.5603&0.5825\\
	HAN&0.6728&0.6318&0.6786 \\
	Longformer&0.6812&0.6425&0.6182 \\
	Att-BLSTM&0.6941&0.6587&0.6910 \\
	BERT&0.7505&0.7260&0.7105 \\
	\hline  
    \textbf{MTT-BERT(w/o KG)}&0.7680&0.7386&0.7714 \\
    \textbf{KG-MTT-BERT}&\textbf{0.8345}&\textbf{0.8054}&\textbf{0.8445} \\
    \hline
  \end{tabular}
\end{table}

\begin{table}
  \label{Experiments on DRG-C Dataset}
  \centering
  \small
  \caption{Experiments on DRG-C Dataset}
  \begin{tabular}{llll}
    \hline
    Method & Acc &F1 &PRAUC \\
    \hline 
    FastText&0.6731&0.6261&0.6827 \\
    Reformer&0.7343&0.7092&0.7673 \\
    Longformer&0.8026&0.7679&0.7931 \\
	TextCNN&0.7961&0.7712&0.8105 \\
	DPCNN&0.8028&0.7775&0.8190 \\
	HAN&0.8240&0.8076&0.8408 \\
	Att-BLSTM&0.8257&0.8050&0.8415 \\
	BERT&0.8372&0.8216&0.8299 \\
	RCNN&0.8415&0.8251&0.8631 \\
	\hline  
    \textbf{MTT-BERT(w/o KG)}&0.8525&0.8267&0.8854 \\
    \textbf{KG-MTT-BERT}&\textbf{0.8825}&\textbf{0.8595}&\textbf{0.9145} \\
    \hline
  \end{tabular}

\end{table}

\paragraph{\textbf{Results Analysis}} In the 5 groups of the 177 groups in the test set of the DRG-A dataset, the comparison of the correct sample numbers of BERT, MTT-BERT, and KG-MTT-BERT is shown in Figure 5.
\begin{figure}[ht]
  \centering
  \label{labels_bar }
  \includegraphics[width=0.99\columnwidth]{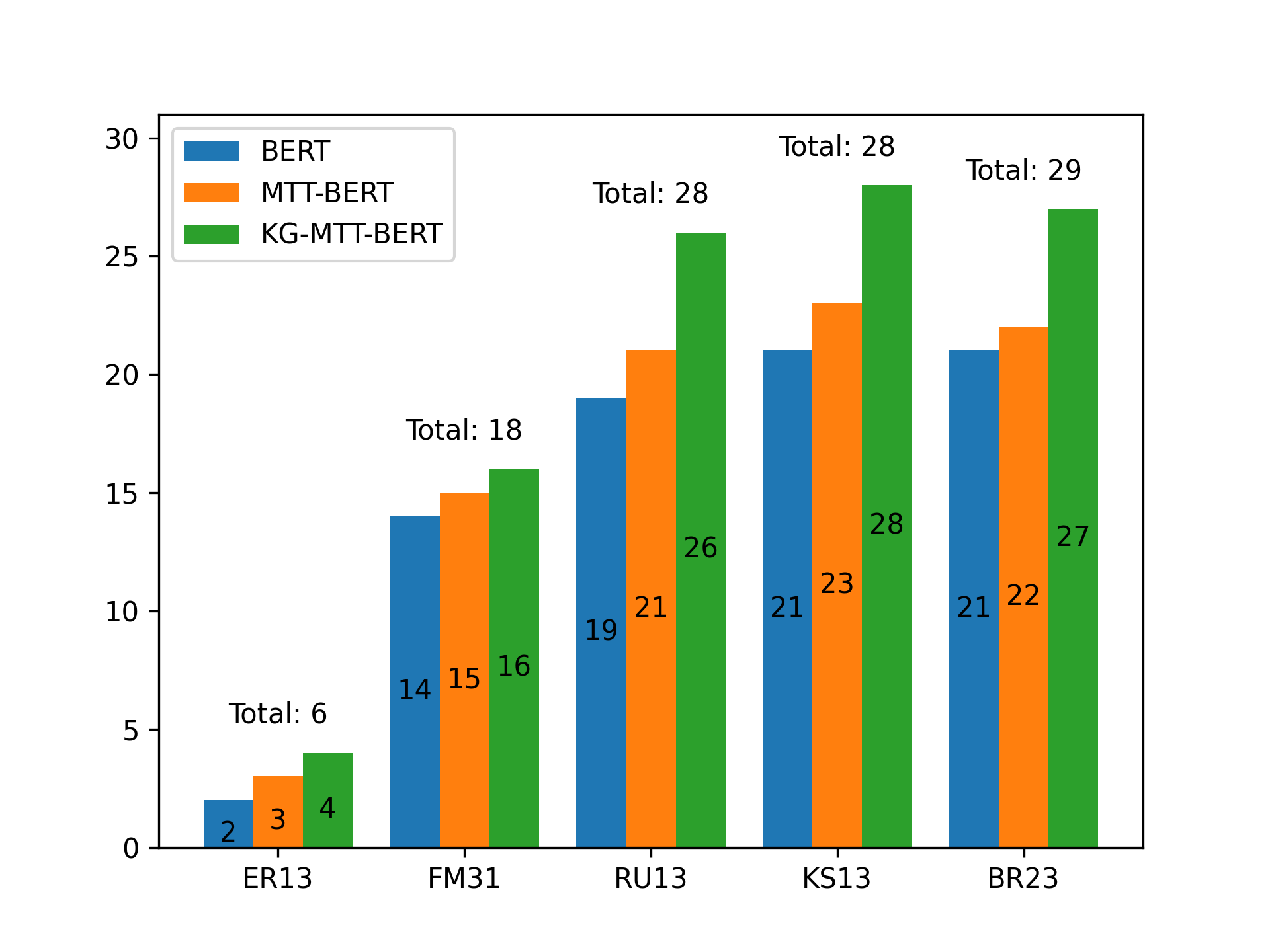}
  \caption{The number of samples that are correctly inferred for 5 categories in the test set of DRG-A dataset.}
\end{figure}

There are 28 samples in the group ``KS13: diabetes, with general complications and comorbidities'' in the test set of the DRG-A dataset. Of these samples, 7 are incorrectly classified by BERT, 5 are incorrect of MTT-BERT, and all samples are correctly inferred by KG-MTT-BERT.

For sample 782 shown in Figure 6, its correct DRG grouping is ``KS13: diabetes, with general complications and comorbidities''. Both BERT and MTT-BERT infer the sample into the group ``BX29: cranial nerve/peripheral nerve disease'' because there are some words related to ``peripheral neuropathy'' in the diagnosis text, hospital course, and history of present illness, such as  ``diabetic peripheral neuropathy'',  ``neurology'',  and ``trophic nerve''. The principal diagnosis ``type 2 diabetes'' is extracted from the diagnosis text, some symptoms related to diabetes are extracted from the chief complaint, such as ``polydipsia'', ``polyuria'', and ``numbness in the limbs''. And the knowledge graph, which includes these entities, is used to enhance the model, so KG-MTT-BERT can make a correct inference. In addition, see Appendix Sample 766 for comparing BERT and MTT-BERT. 
\begin{figure}[ht]
  \centering
  \label{simple_782}
  \includegraphics[width=0.99\columnwidth]{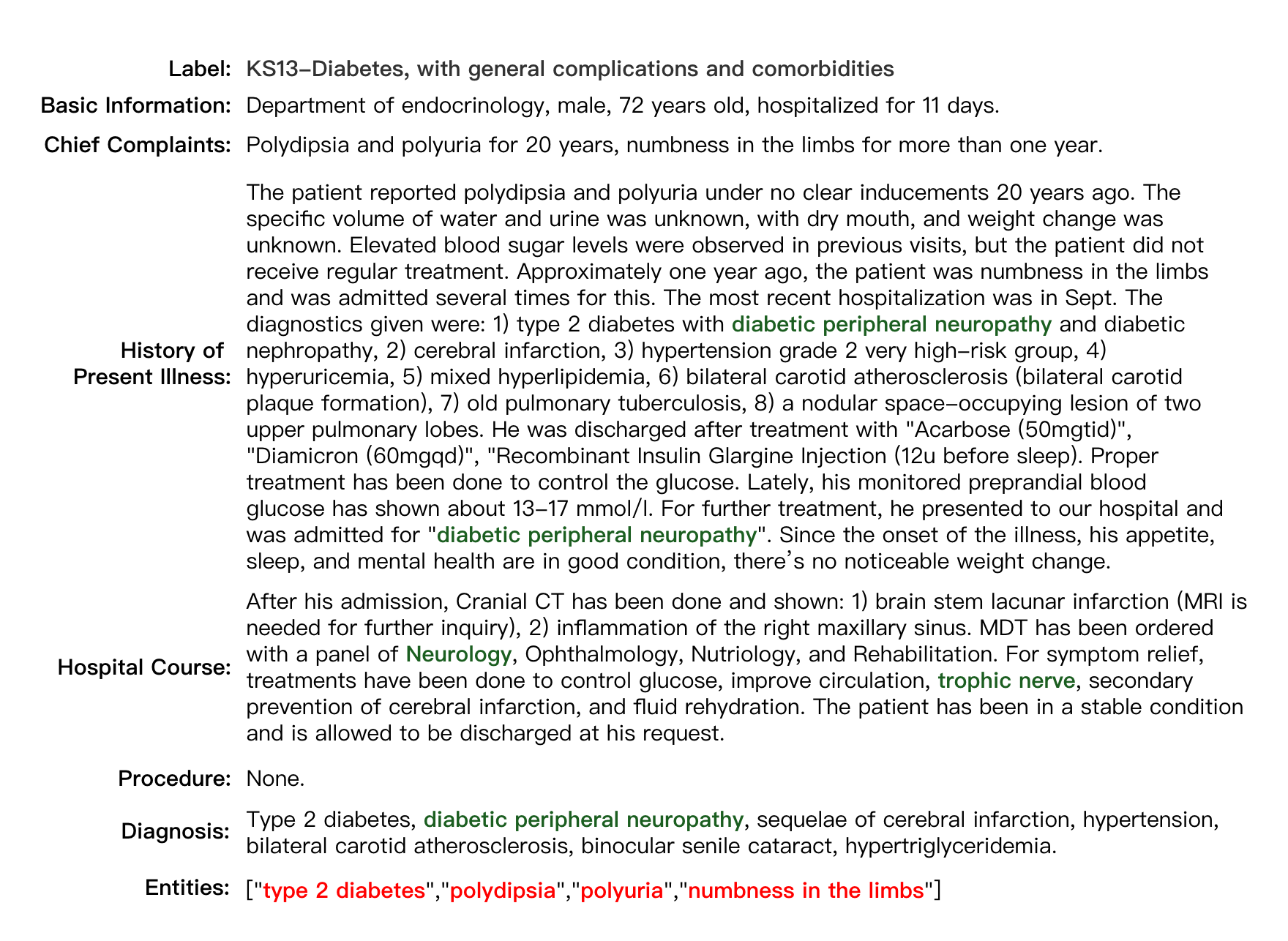}
  \caption{Sample 782 in the test set of the DRG-A dataset (translated from Chinese health record texts).}
\end{figure}

\subsection{Hyperparameters}
In this subsection, we study the effect of the hyperparameters on our model. Due to the limited space, we use the DRG classification task on the DRG-A dataset as an example. We focus on three hyperparameter sets including batch size \textit{BS}, learning rate \textit{LR}, and maximum text length \textit{ML}. We vary the values in each hyperparameter set with other sets fixed. The results for different hyperparameter settings are shown in Table 5. We can observe that the varying hyperparameter \textit{BS} and \textit{LR} have little effect on performance with less than 0.02 changes of all metrics. However, the performance drops dramatically as the \textit{ML} decreases, especially when \textit{ML} is less than 64. This indicates that including full text for modeling is important for text classification. 

\begin{table}[htbp]
  \label{Experiments on Varying Hyperparameters}
  \centering
  \small
  \caption{Experiments on Varying Hyperparameters}
  \begin{tabular}{llll}
    \hline
    Hyper Parameters & Acc & F1 &PRAUC \\
    \hline  
    \multicolumn{4}{l}{Varying BS(LR=1e-5, ML=512)}\\
    \hline  
    BS=2 &\textbf{0.9256}&\textbf{0.9091}&0.9303 \\
	BS=4 &0.9239&0.9040&0.9307 \\
	BS=8 &0.9179&0.8961&\textbf{0.9310} \\
	\hline  
	\multicolumn{4}{l}{Varying
	LR(BS=8, ML=512)}\\
	\hline  
	LR=5e-6 &0.9171&0.8897&0.9269 \\
	LR=1e-5 &0.9179&0.8961&0.9310 \\
	LR=2e-5 &\textbf{0.9247}&\textbf{0.9044}&\textbf{0.9366} \\
	\hline
	\multicolumn{4}{l}{Varying ML(BS=8, LR=1e-5)}\\
    \hline  
    ML=512 &0.9179&0.8961&\textbf{0.9310} \\
	ML=128 &\textbf{0.9226}&\textbf{0.9002}&0.9280 \\
	ML=64 &0.9137&0.8915&0.9266 \\
	ML=32 &0.8813&0.8540&0.8851 \\
	ML=16 &0.8307&0.7882&0.8459 \\
	ML=8 &0.7567&0.7260&0.7947 \\
	\hline  
  \end{tabular}
\end{table}

\subsection{Ablation Study}
We have demonstrated strong empirical results for our model in the DRG classification task. In this section, we perform various ablation studies in different aspects of our model to get a better understanding of their relative importance. The effects of encoding methods, pooling strategies, and knowledge graph are shown in Table 6. Due to the limited space, we only report the comparison in the DRG-A dataset. 

\paragraph{\textbf{Encoding Method}} We compare text-level encoding and token-level encoding with the global max pooling strategy and omission of the knowledge graph. The token-level encoding can achieve better performance than text-level encoding with greater than 0.016 improvement of all metrics, indicating that fine-grained encoding is better than coarse-grained encoding for this task. 

\paragraph{\textbf{Pooling Strategy}} Various pooling strategies have been investigated with token-level encoding and omission of the knowledge graph. The global max pooling strategy can achieve the best performance among all the strategies. The global max pooling is slightly better than the simple transformer layer in all metrics. The DRG is mainly determined by the diagnoses and procedures, which are related to a small portion of all input texts that play a key role in inference. There is no surprise that the simple average is the worst since there is all token are treated equally with no information attended. In addition, the comparison of vector-level versus value-level contextual attention indicates that the weight assignment at the fine-grained level shows significant improvement. 

\paragraph{\textbf{Knowledge Graph}} We compare our model with and without knowledge graph as in the last part of Table 6. The inclusion of the knowledge graph can improve the performance significantly. This indicates that the inclusion of a knowledge graph for medical domain applications is crucial for its performance. 

\begin{table}[ht]
  \label{Ablation Study on DRG-A Dataset}
  \centering
  \small
  \caption{Ablation Study on DRG-A Dataset}
  \begin{tabular}{llll}
    \hline
	Encoding(w/o KG) & Acc & F1 &PRAUC \\
    \hline 
    Text-Level  &0.8545&0.8190&0.8524 \\
    Token-Level &\textbf{0.8703}&\textbf{0.8425}&\textbf{0.8741} \\    
    \hline
    Pooling(w/o KG) & Acc & F1 &PRAUC \\
    \hline 
    Average/Vector-Level &0.5742&0.5124&0.5809 \\
	Contextual/Vector-Level &0.8214&0.7725&0.7809 \\
	Contextual/Value-Level &0.8486&0.8131&0.8256 \\
	Transformer/Vector-Level &0.8664&0.8384&0.8646 \\
	Max &\textbf{0.8703}&\textbf{0.8425}&\textbf{0.8741}  \\
	\hline 
	using KG or not & Acc & F1 &PRAUC \\
    \hline 
    MTT-BERT(w/o KG) &0.8703&0.8425&0.8741 \\
    KG-MTT-BERT &\textbf{0.9179}&\textbf{0.8961}&\textbf{0.9310} \\
	\hline
  \end{tabular}
\end{table}

\paragraph{\textbf{Case Study}} We demonstrate an example of DRG category GC25 (anus and gastrointestinal stoma surgery) and the six types of input texts are shown in Figure 7. The most important words selected by the global max pooling layer are marked in red. With basic medical knowledge and common sense, the marked words are all related to the category of this sample and many irrelevant words are indeed not selected. 

\begin{figure}[ht]
  \centering
  \label{important words }
  \includegraphics[width=0.99\columnwidth]{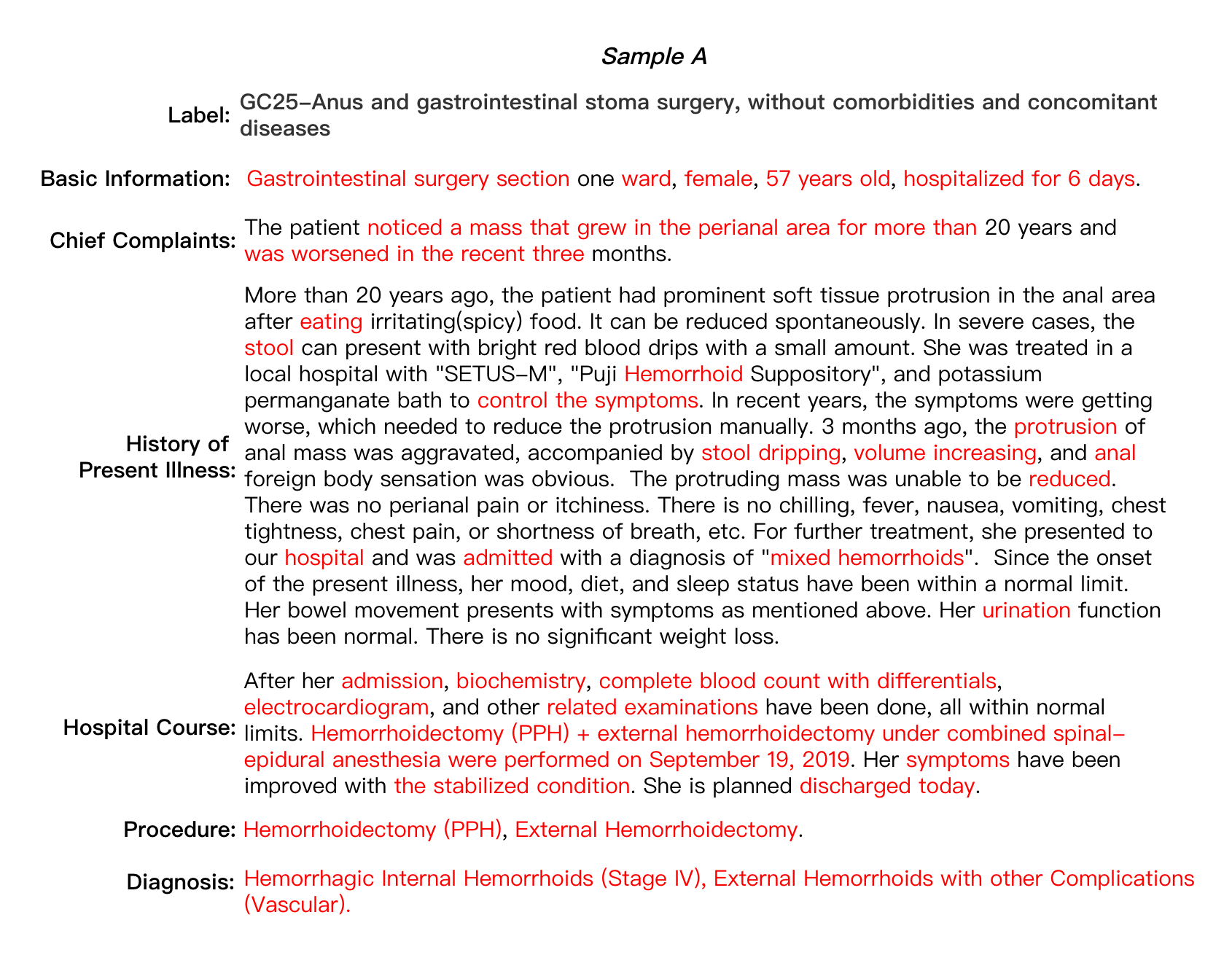}
  \caption{The important words selected by the global max pooling layer. The selected words are marked in red. It is translated from the Chinese health record in Figure 8 in Appendix A.1.}
  
\end{figure}

\section{Conclusions}
In this work, we propose a novel architecture KG-MTT-BERT for multi-type text classification. The new model can deal with different types of texts and integrate domain knowledge. We evaluate our model on the DRG classification task of three medical datasets with superior performance comparing to other baselines and state-of-the-art models. The ablation studies investigate the impact of different encoding methods, pooling strategies, and the integration of domain knowledge. In the future, we will investigate deep learning methods to couple the predicted DRG with the resource management model to allocate scarce hospital resources.

\bibliographystyle{ACM-Reference-Format}
\bibliography{paper}


\clearpage
\appendix
\section{Appendix}
\subsection{Multi-Type Medical Text}
\label{sec:multi_type_text_appendix}
We view different fields in EHR as different types due to different purposes and diverse text generation processes. The chief complaint is the subjective description from the patient and is mostly within 20 Chinese characters. The history of the present illness is a lengthy detailed objective observation written by the doctor at the admission of the patient. The hospital course is a summary of the current visit and most of the text is generated by a doctor-customized template or copied from standard reports then edited by the doctor. Diagnosis and procedure fields are the free text of diagnosis and procedure mixed with standard ICD names and non-standard names. Basic information is from tabular data such as age and gender. Therefore, the text in the different fields could be objective or subjective, wrote by the doctor or generated using a template, and free text or tabular data. The diversity of medical text is interesting to investigate more in the future. In Figure 8, we represent a sample with six types of medical texts.

\begin{figure}[ht]
  \centering
  \label{multi-type medical text}
  \includegraphics[width=1\columnwidth]{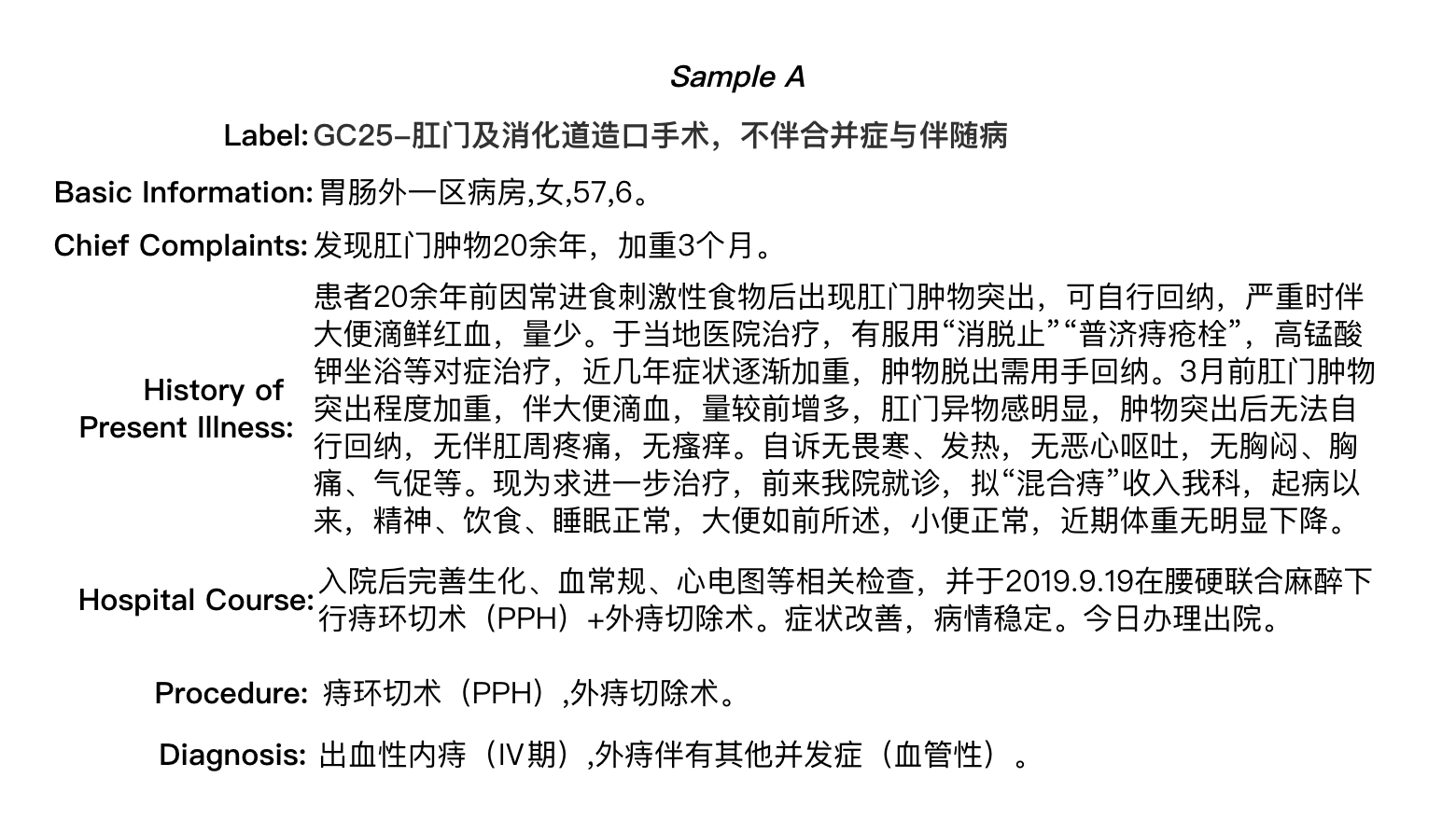}
  \caption{A sample with six types of medical texts.}
\end{figure}

\subsection{Hospital Classification}
\label{sec:hc_appendix}
According to ``the hospital classification system'' of the Ministry of Health of the People's Republic of China, both public and private hospitals in China are classified as Primary (Class I), Secondary (Class II), or Tertiary (Class III) institutions based on their ability to provide medical care, medical education, and conduct medical research. A primary hospital is typically a township hospital that provides preventive care, minimal health care, and rehabilitation services. A secondary hospital is defined as a regional hospital that provides comprehensive medical care, medical education, and medical research for the region. A tertiary hospital is defined as cross-regional, providing comprehensive and specialized medical care with a high level of medical education and scientific research.

\subsection{Medical Knowledge Graph}
\label{sec:medical_knowledge_graph_appendix}
We build the Knowledge Graph for each dataset independently using entities extracted from each training set.

\paragraph{\textbf{Entity Set}} There are four types of entities: symptom, diagnosis, procedure, and DRG category. The symptom set is from the Chinese medial knowledge database OMAHA \footnote[3]{https://www.omaha.org.cn/} with 74963 symptom entities. Diagnosis and procedure are from ICD-10 and ICD-9-CM-3, respectively. The number of DRG categories for DRG-A, DRG-B, and DRG-C datasets are 804, 783, and 618, respectively.

\paragraph{\textbf{Entity and Relation Extraction}} We extract the entity from the medical text by going through the NER and the subsequence matching algorithm, and these extracted entities are matched to the standard entity names through the string similarity measure algorithm based on the LCS (longest common subsequence). The symptoms are extracted from the EHR field chief complaint and history of present illness. The principal diagnosis and principal procedure entities are extracted from EHR field diagnosis and procedure (free text), respectively. DRG is the category label. The extracted entities are linked if they are in the same hospital visit (co-occurrence relation). 

\paragraph{\textbf{Knowledge Graph Statistics}} For the DRG-A dataset, entity statistics are DRG (767), diagnosis (2615), procedure (1092), and symptom (617). Relation statistics are DRG-diagnosis (4041), DRG-procedure (1749), diagnosis-procedure (2967), diagnosis-symptom (806). 

For the DRG-B dataset, entity statistics are DRG (745), diagnosis (5260), procedure (2152), and symptom (1057). Relation statistics are DRG-diagnosis (8099), DRG-procedure (3252), diagnosis-procedure (6801), diagnosis-symptom (1214). 

For the DRG-C dataset, entity statistics are DRG (515), diagnosis (1514), procedure (753), and symptom(435). Relation statistics are DRG-diagnosis (2229), DRG-procedure (1035), diagnosis-procedure (1606), diagnosis-symptom (557).

We want to note that the symptom set is the same in the three datasets. Diagnosis, procedure, and DRG entity sets are different across three datasets. In China, ICD has more than 20 local versions with different translation and code expansion from WHO. DRG has five more versions.

\subsection{Pseudo-code}
\label{sec:pcode_appendix}
The pseudo-code of KG-MTT-BERT is as \textbf{Algorithm 1}, which applies the token-level encoder and the global max pooling. We will open our code to GitHub, and the approval process for opening the three datasets is in progress. And all of these datasets are desensitized and used under the license from the data provider.

\begin{algorithm}[htb]
    \small
    \caption{KG-MTT-BERT}
    \begin{algorithmic}[1]
        \Require Hyperparameters: $BS$, $LR$, and $ML$; Knowledge Graph: $KG$; Texts: $Xs$; Entities: $T$; Labels: $y$.
        \Ensure Accuracy, F1-score, PRAUC
        \Procedure{training}{}
            \State Apply the model KG2E to train the KG to obtain the embedding matrix $E^{KG}$, which is used as the initial embedding matrix of all entities.
            \State Load the model ``chinese\_L-12\_H-768\_A-12'', and initialize the BERT Encoder and Transformer (Entity Encoder). 
            \For{$i = 1 \to Epoch$}  
                \For{$batch_j$ in training set}
                        \State $[X_1, X_2, X_3, X_4, X_5, X_6, T] \gets batch_j$
                        \State $H_k \gets BERT Encoder(X_k), k \in [1,...,6]$
                        \State $M^{KG} \gets Entity Encoder(T)$
                        \State $C \gets Concat(H_1, H_2, H_3, H_4, H_5, H_6)$
                        \State $e \gets GlobalMaxPooling(C)$
                        \State $e^{KG} \gets GlobalMaxPooling(M^{KG})$
                        \State $\hat{y} \gets classify(concat(e, e^{KG}))$
                        \State compute loss $<y, \hat{y}>$
                        \State update the network parameters basis on loss using Back Propagation 
                \EndFor  
                \State compute Accuracy, F1-score, PRAUC on validation set
            \EndFor
            \State $\mathrm{best\_model} \gets \mathop{\arg\max}_{model}\{\mathrm{Accuracy}\}$
            \State \Return{best\_model}
        \EndProcedure
        \Procedure{evaluation}{}
            \State compute Accuracy, F1-score, PRAUC on test set
            \State \Return{Accuracy, F1-score, PRAUC}
        \EndProcedure
    \end{algorithmic}
\end{algorithm}

\subsection{Execution Information}
\label{sec:exe_appendix}
All baseline algorithms are executed with multiple sets of hyperparameters, and each set of hyperparameters will be executed 5 times to get the results, and then the average of the results of 5 times of the best hyperparameter set is calculated as the performance of the algorithm.

For Fasttext, DPCNN, TextCNN, Att-BLSTM, RCNN, and HAN, the main hyperparameters and search space are as follows: the learning rate: \{1e-2, 5e-3, 2e-3, 1e-3\} and the batch size: \{4, 8, 16\}.

For Reformer, the model parameter is initialized as follows: BERT-Chinese-base used for the same layers as in BERT and a truncated normal distribution used for other layers. The hyperparameters and search space are: the maximum length of text: \{1024, 1408, 2048\}, the learning rate: \{1e-5, 2e-5, 5e-6\}, the attention method of each layer: \{``local,lsh,local,lsh,local,lsh,local,lsh,local,lsh,local,lsh'', ``local,local,lsh,local,local,local,lsh,local,local,local,lsh,local'' \}, the number of buckets: \{16, 64, 128\}, and the batch size of training: \{4, 8\}.

For Longformer, we experiment with two types of model parameter initialization methods: 1) BERT-Chinese-base used for the same layers as in BERT and a truncated normal distribution used for other layers; 2) the pre-trained Longformer-Chinese-base as the initial parameters for the model. The hyperparameters and search space are: the maximum length of the text: \{1024, 1408, 2048\}, the learning rate: \{1e-5, 2e-5, 5e-6\}, the size of attention window: \{64, 128\}, and the batch size of training: \{4, 8\}. 

For BERT, we use BERT-Chinese-base for initialization. The main hyperparameters and search space are as follows: the learning rate: \{1e-5, 2e-5, 5e-6\}, the batch size: \{4, 8\}, and the maximum text length: \{128, 256, 512\}. 

The number of encoder layers is 12 for Reformer, Longformer, and BERT. The training epoch is 50 for these baseline algorithms. The input texts are concatenated into a long text in the order of chief complaint, procedure, diagnosis, hospital course, history of present illness, and basic information, which is based on the importance of each text type for the DRG classification task. In the maximum length of text hyperparameters, we include the value 1408, which is the sum of the maximum length for each field text set in our model: 64+128+128+512+512+64=1408.

The number of parameters in KG-MTT-BERT is about 113.5 million. The hyperparameters of our model are batch size \textit{BH}, learning rate \textit{LR}, and maximum text length \textit{ML}. We chose these hyperparameter values based on experience and related works. We study the effect of the hyperparameters on KG-MTT-BERT in subsection Hyperparameters. The training epoch is 30 for our model and is 50 for the rest of the models. The average runtime of our model is about 0.39s for batch training and is 0.14s for batch inference (\textit{BH}=8, \textit{ML}=512). 

\subsection{Sample 766}
\label{sec:another_sample}
For sample 776 presented in Figure 9, the correct group is ``KS13: diabetes, with general complications and comorbidities'', which BERT infers the sample into the group ``BX29: cranial nerve/peripheral nerve disease''. BERT concatenates the six input texts without differentiating their importance. The input texts are concatenated as one long text in the order of chief complaint, procedure, diagnosis, hospital course, history of present illness, and basic information. The total length of these texts is 1257 (in Chinese), and the length of the text before the history of present illness is 930 (in Chinese), so the history of present illness is truncated completely. However, the history of present illness (310 tokens in Chinese) are all related to ``diabetes'', which plays a key role in the grouping of the sample. Additionally, there are some words related to the `BX29'' group in the retained text, such as ``peripheral neuropathy'' and ``trophic nerve''. For these reasons, BERT cannot make a
proper inference. However, MTT-BERT treats each type of text separately so that the sample is correctly classified.

\begin{figure}[ht]
  \centering
  \label{sample 766}
  \includegraphics[width=1\columnwidth]{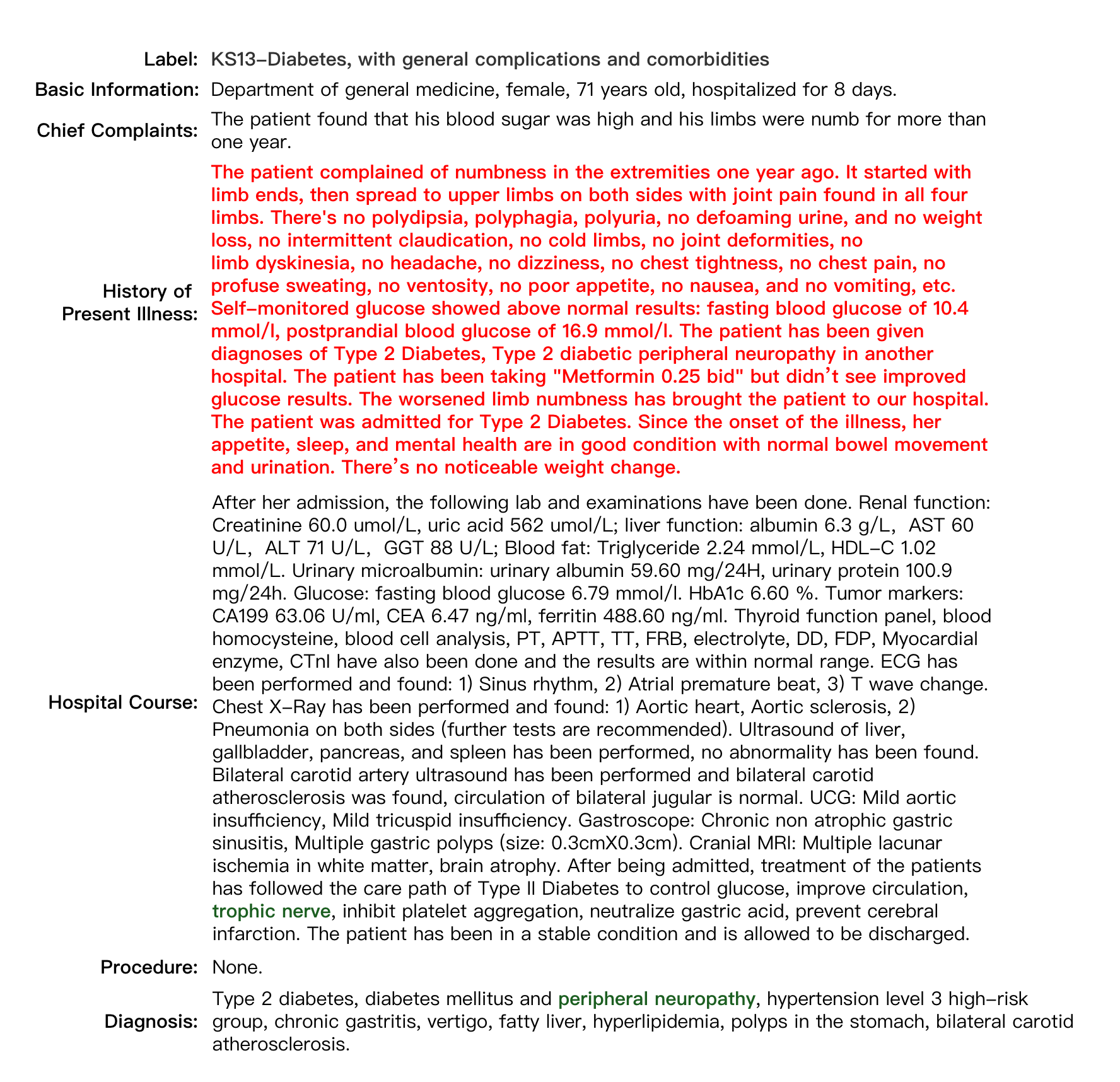}
  \caption{Sample 766 in the test set of the DRG-A dataset (translated from Chinese health record texts). }
\end{figure}

\end{document}